\documentclass[letterpaper, 10 pt, conference]{ieeeconf}  % Comment this line out if you need a4paper

\IEEEoverridecommandlockouts                              % This command is only needed if
\usepackage{graphics} % for pdf, bitmapped graphics files
\usepackage{amsmath} % assumes amsmath package installed
\usepackage{amssymb}  % assumes amsmath package installed
\usepackage{amssymb}

\usepackage{hyperref}
\usepackage{url}
\usepackage{graphicx}
\usepackage{algorithm}%
\usepackage{algorithmicx}%
\usepackage{algpseudocode}%
\usepackage{bm}
\usepackage{ragged2e}
\usepackage[table]{xcolor}
\usepackage{booktabs}

\newcounter{remark}
\newenvironment{remark}{
\refstepcounter{remark}
\par\medskip
\noindent\textbf{Remark~\theremark. }\rmfamily
}{\medskip}

\newcounter{assumption}
\newenvironment{assumption}{
\refstepcounter{assumption}
\par\medskip
\noindent\textbf{Assumption~\theassumption. }\rmfamily
}{\medskip}

\title{\LARGE \bf
Taming the Adversary: A Cost-to-Disturbance Ratio Approach to Adversarial Reinforcement Learning
}

\author{Taeho Lee and Donghwan Lee% <-this % stops a space
\thanks{This work was supported by the National Research Foundation of Korea (NRF) grant funded by the Korea government (MSIT) (No. RS-2026-25510069), and by the BK21 FOUR program funded by the Ministry of Education (Republic of Korea).}% <-this % stops a space
\thanks{Taeho Lee and Donghwan Lee are with the School of Electrical Engineering, Korea Advanced Institute of Science and Technology (KAIST), Daejeon 34141, Republic of Korea.
        {\tt\small eho0228@kaist.ac.kr, donghwan@kaist.ac.kr}}%
}

\begin{document}

\maketitle
\thispagestyle{empty}
\pagestyle{empty}

%%%%%%%%%%%%%%%%%%%%%%%%%%%%%%%%%%%%%%%%%%%%%%%%%%%%%%%%%%%%%%%%%%%%%%%%%%%%%%%%
\begin{abstract}

Reinforcement learning (RL) policies trained in simulation often degrade once deployed on real systems, where the controller must reject external disturbances that were never encountered in simulation. Robust RL addresses this by exposing the controller to perturbations while it learns, through domain randomization, adversarial minimax formulations, or probabilistic mixtures of protagonist and adversarial behavior. However, an unregulated disturbance mechanism destabilizes training and often collapses nominal performance relative to standard, non-robust methods.

We propose cost-to-disturbance ratio adversarial training (CoDRA), a framework that expresses the controller--adversary trade-off as a ratio of accumulated cost to accumulated squared disturbance norm, and optimizes it through a self-normalized actor--critic update. In this algorithm, each value term is scaled by a stop-gradient normalization constant computed from the current batch. This moderates the adversary's incentive without altering the controller's own update, and requires neither an explicit disturbance penalty nor an auxiliary trade-off parameter.

We evaluate CoDRA on two MuJoCo pendulum environments under force and mass sweeps. On InvertedDoublePendulum, CoDRA attains the lowest cost at every force level, including a force outside the range seen during training, and in all but one cell of the mass grid, whereas its advantage is less pronounced on the milder InvertedPendulum.

\end{abstract}

%%%%%%%%%%%%%%%%%%%%%%%%%%%%%%%%%%%%%%%%%%%%%%%%%%%%%%%%%%%%%%%%%%%%%%%%%%%%%%%%
\section{Introduction}

In recent years, reinforcement learning (RL) agents have achieved strong performance in complex, high-dimensional environments such as competitive games~\cite{RL1,DQN} and nonlinear control systems~\cite{DDPG,RL5}. Despite these successes, RL agents remain highly sensitive to conditions that differ from those seen during training~\cite{AdvRLSurvey,RARL,RAP,RARLineq}. Physical systems are continually subject to external disturbances---wind, contact forces, payload shifts, and actuator bias---as well as unmodeled dynamics and parameter variations that simulation rarely reproduces~\cite{SalvatoRealityGap}. A policy that performs well in simulation can therefore degrade, or lose stability altogether, once deployed, which is a serious concern in safety-critical domains such as robotics, autonomous systems, and industrial control~\cite{GarciaFernandezSafeRL}. Limiting the influence of such disturbances on closed-loop performance is therefore a central requirement for practical RL-based control, particularly for sim-to-real transfer~\cite{ZhaoSimToRealSurvey}.

Adversarial RL addresses this problem by introducing a second agent that perturbs the controller during training, either by applying external forces to the system~\cite{AdvRLSurvey,RARL,lee2025robust} or by perturbing the controller's actions~\cite{ARDDPG}. The adversary injects a disturbance that challenges the controller, while the controller attempts to maintain performance under it, so that the learned policy is exposed during training to disturbances actively chosen to be harmful rather than to nominal conditions alone. However, direct adversarial training often becomes unstable: an unregulated adversary can produce disturbances that are excessively large, and these disturbances can dominate the optimization process. As a result, policy improvement becomes difficult.

To overcome these limitations, we propose cost-to-disturbance ratio adversarial training (CoDRA), a framework for learning policies that remain robust to external disturbances in continuous control tasks. The controller is trained alongside an adversarial disturbance policy that perturbs it during training. To stabilize this interaction, we introduce a self-normalized actor-critic update in which the value terms used by the controller and the adversary are each scaled by a stop-gradient normalization constant estimated from the current batch. This regulates the adversary's incentive so that it can still meaningfully challenge the controller without being rewarded for pushing the disturbance to extremes. Although the adversary acts only through a force disturbance channel during training, we find that the resulting policy also transfers to parametric mass mismatch that it never encountered during training---though this transfer is not uniform across tasks.

\begin{itemize}
\item We formulate adversarial training around the ratio of accumulated cost to accumulated squared disturbance norm, and show that a logarithmic transformation of this ratio yields a self-normalized actor-critic update in which each agent's value term is scaled by a stop-gradient normalization constant. This regulates the adversary's disturbance magnitude without an explicit penalty term or auxiliary trade-off parameter.
\item We implement this update within the deep deterministic policy gradient (DDPG)~\cite{DDPG} framework. The result is a practical actor-critic instantiation of CoDRA that uses only single-critic value estimates for both the controller and the adversary.
\item We show that, on InvertedDoublePendulum, where a fall terminates the episode with a penalty, CoDRA attains the lowest cost among the compared methods at every force level, including a force outside the range the adversary was allowed during training, and in $8$ of the $9$ mass-grid cells; on the milder InvertedPendulum, which has no early termination, its advantage is less pronounced. The benefit therefore appears largest where an unregulated adversary is most destabilizing.
\end{itemize}

\section{Related Works}
\label{sec:relatedworks}
Robust RL methods expose the controller to perturbations during training. Adversarial training injects disturbances as external forces~\cite{RARL} or by perturbing the action~\cite{ARDDPG}, whereas domain randomization perturbs environment parameters~\cite{SalvatoRealityGap,ZhaoSimToRealSurvey,Tobin2017DR,GarciaFernandezSafeRL}. We focus on the former, in which an adversary applies an external force that the controller must reject. Since force-space adversarial training has also been reported to improve robustness to parametric perturbations~\cite{RARL}, we use domain randomization as a non-adversarial reference and evaluate robustness to both force disturbances and mass mismatch.

Robust RL~\cite{RRL} traces back to $H_\infty$-style differential games~\cite{Hbook}, which give strong guarantees but are intractable beyond low-dimensional systems. Deep-RL extensions~\cite{RARL,RARLineq,ARDDPG,HRL3} scale this to high-dimensional control: robust adversarial reinforcement learning (RARL)~\cite{RARL} casts it as a two-player zero-sum game, but the adversary can converge faster than the protagonist, which produces overly aggressive disturbances and unstable training~\cite{RAP,RARLineq,HRL3}. Subsequent work imports stability constraints from robust control---dissipativity/$L_2$-gain conditions~\cite{RARLineq} or $H_\infty$-inspired regularization~\cite{HRL3}---at the cost of extra tuning and overhead, or restricts where the adversary can act, as in action-robust Markov decision processes (MDPs) that confine it to perturbing or replacing the action~\cite{ARDDPG,RRLforquad}, which bounds its influence by construction but suits action uncertainty more than persistent dynamics disturbances. Domain randomization~\cite{Tobin2017DR} instead randomizes physical parameters directly and needs no adversary, but its effectiveness depends on the chosen randomization ranges.

Most closely related is our earlier framework~\cite{lee2025robust}, which also casts disturbance attenuation as an $H_\infty$-style two-player game, but regulates the adversary by adding a quadratic disturbance penalty to the reward and adapting its weight online. The present work leaves the reward unchanged and regulates the adversary through the objective itself.

In contrast to all of these, we neither modify what the adversary perturbs nor add an explicit disturbance penalty: we express the objective as a ratio of accumulated cost to accumulated squared disturbance norm, and stabilize its optimization through a self-normalized value update that requires no auxiliary trade-off parameter or explicit stability constraint. Our experiments indicate that this improves robustness to force disturbances and transfers to parametric mass mismatch, most clearly on InvertedDoublePendulum, without action-space restriction or parameter randomization.

\section{Preliminaries}

\subsection{Two-player zero-sum Markov game}
\label{sec:AdvRL}
In a two-player zero-sum Markov game (TZMG)~\cite{perolat2015approximate}, a controller and an adversary share a state space $\mathcal{S}$ and have action spaces $\mathcal{A}$ and $\mathcal{W}$, respectively. At the current state $s\in\mathcal{S}$, the controller selects an action $a\in\mathcal{A}$ and the adversary simultaneously selects a disturbance $w\in\mathcal{W}$; the state then transitions to $s'\in\mathcal{S}$ with probability $P(s'|s,a,w)$, incurring a cost $c(s,a,w,s')$, where $c:\mathcal{S}\times\mathcal{A}\times\mathcal{W}\times\mathcal{S}\rightarrow\mathbb{R}$ is the cost function. For convenience, we consider a deterministic cost function and simply write $c_{k+1} := c(s_k, a_k, w_k, s_{k+1})$, where $k\in \{0,1,...\}$ is the time step.
Let $\pi: \mathcal{S} \rightarrow \mathcal{A}$ and $\mu: \mathcal{S} \rightarrow \mathcal{W}$ denote the policies of the controller and the adversary, respectively. The objectives of the controller and the adversary are to minimize and maximize, respectively, the cumulative discounted cost over an infinite horizon, $J^{\pi,\mu}=\mathbb{E}\left[ {\left.\sum_{k=0}^{\infty}\gamma^k c_{k+1}\right|\pi,\mu}\right]$, where $\gamma\in[0,1)$ is the discount factor, $(s_0,a_0,w_0,s_1,a_1,w_1,...)$ is a state-action trajectory generated by the Markov chain under policies $\pi$ and $\mu$, and $\mathbb{E}[\cdot|\pi,\mu]$ is an expectation conditioned on the policies $\pi$ and $\mu$.

Specifically, we consider a TZMG with state transition dynamics given by
\begin{align*}
    s_{k+1} \sim P(\cdot|s_k, a_k, w_k),
\end{align*}
where $s_k$ and $s_{k+1} \in \mathbb{R}^n$ denote the current and next states, respectively, $a_k \in \mathbb{R}^m$ is the action selected by the controller, and $w_k \in \mathbb{R}^d$ represents the disturbance generated by the adversary.
At each time step $k$, both players select their decisions based on the current state $s_k$ through deterministic policies.
Accordingly, we define deterministic state-feedback policies for the controller and the adversary as $\pi_\theta: \mathbb{R}^n \rightarrow \mathbb{R}^m$ and $\mu_\phi: \mathbb{R}^n \rightarrow \mathbb{R}^d$, which are parameterized by $\theta$ and $\phi$, respectively,
\begin{align*}
    a_k = \pi_{\theta}(s_k),  \quad w_k = \mu_{\phi}(s_k).
\end{align*}
Under these state-feedback policies, the closed-loop system dynamics reduce to a Markov chain governed by
\begin{align*}
s_{k+1} \sim P\bigl(\cdot|s_k, \pi_\theta(s_k), \mu_\phi(s_k)\bigr).
\end{align*}

\begin{figure*}[ht]
    \centering
    \includegraphics[width=0.7\linewidth]{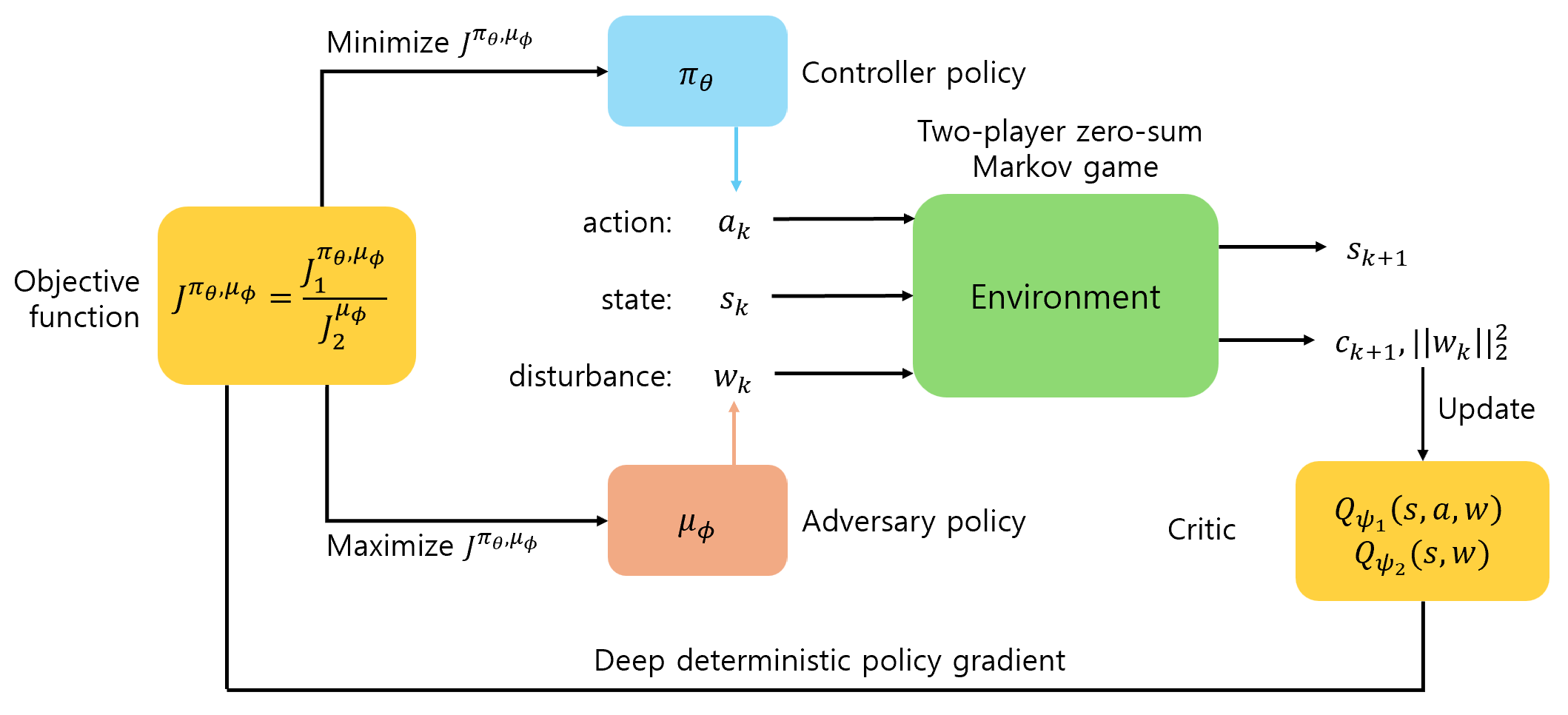}
    \caption{Overview of cost-to-disturbance ratio adversarial training (CoDRA). The controller and the adversary interact with the environment, generating the action $a_k$ and the disturbance $w_k$ from the state $s_k$. The critics $Q_{\psi_1}(s,a,w)$ and $Q_{\psi_2}(s,w)$ are trained using $c_{k+1}$ and $\|w_k\|_2^2$, respectively. The policy of controller $\pi_\theta$ is updated to minimize the self-normalized objective derived from $J^{\pi_\theta,\mu_\phi}$ while the policy of adversary $\mu_\phi$ is updated to maximize it.}
    \label{fig:CoDRA}
\end{figure*}

\section{Robustness Objective}
In TZMG, the controller seeks a control policy that minimizes the expected cumulative cost, whereas the adversary aims to maximize it by choosing disturbances. Under deterministic policies $\pi_\theta$ and $\mu_\phi$, we define the primary performance objective as the expected discounted cumulative cost
\begin{align}
    J_1^{\pi_\theta,\mu_\phi}
    := \mathbb{E}\!\left[\left.\sum_{k=0}^{\infty}\gamma^k c_{k+1}\ \right| \pi_\theta,\mu_\phi\right].
    \label{eq:J1}
\end{align}
Accordingly, the robust control problem in TZMG can be formulated as the controller minimizing and the adversary maximizing $J_1^{\pi_\theta,\mu_\phi}$. Directly optimizing this objective, however, often leads to unstable learning dynamics: since the adversary seeks to maximize the cumulative cost, it can increase the objective arbitrarily by enlarging the disturbance magnitude in the absence of explicit regularization. This behavior destabilizes learning and prevents the actor-critic updates from converging to a meaningful policy pair.

To mitigate this issue, we introduce an additional quantity that measures the cumulative squared disturbance norm,
\begin{align}
    J_2^{\mu_\phi}
    := \mathbb{E}\!\left[\left.\sum_{k=0}^{\infty}\gamma^k \|w_k\|_2^2\ \right| \mu_\phi\right].
\end{align}
Together, $J_1^{\pi_\theta,\mu_\phi}$ and $J_2^{\mu_\phi}$ suggest the ratio
\begin{align}
    \label{eq:J}
    J^{\pi_\theta,\mu_\phi}
    = \frac{J_1^{\pi_\theta,\mu_\phi}}{J_2^{\mu_\phi}}
    = \frac{
        \mathbb{E}\!\left[ \sum_{k=0}^{\infty} \gamma^k c_{k+1} \,\middle|\, \pi_\theta, \mu_\phi \right]
    }{
        \mathbb{E}\!\left[ \sum_{k=0}^{\infty} \gamma^k \| w_k \|_2^2 \,\middle|\, \mu_\phi \right]
    }
\end{align}
as a natural candidate objective, directly capturing the trade-off between task cost and disturbance magnitude.

\begin{remark}
This ratio objective~\eqref{eq:J} is motivated by $H_\infty$ control, where the performance criterion is expressed as a ratio of output energy to disturbance energy. In particular, the $H_\infty$ norm, $\|T_\pi\|_\infty$, characterizes the worst-case disturbance-to-output gain,
\begin{align*}
\|T_\pi\|_\infty^2
=
\sup_{w \neq 0}
\frac{\sum_{k=0}^{\infty} \|y_k\|_2^2}
     {\sum_{k=0}^{\infty} \|w_k\|_2^2},
\end{align*}
where $y_k$ is the performance output and $T_\pi$ is the closed-loop map from $w$ to $y$.
We emphasize that this analogy is motivational: $J^{\pi_\theta,\mu_\phi}$ is a fixed-policy ratio evaluated under $(\pi_\theta,\mu_\phi)$, not a supremum over adversarial disturbances, and it does not constitute a formal $H_\infty$-type robustness guarantee. Establishing such formal guarantees is an important direction for future work.
\end{remark}

\section{Actor and Critic Update}
\label{sec:algorithm}
We now derive the actor-critic updates from the gradient of the ratio $J^{\pi_\theta,\mu_\phi}$ in~\eqref{eq:J} with respect to the policy parameters $\theta$ and $\phi$.

Differentiating the ratio~\eqref{eq:J} directly with the quotient rule yields a gradient estimator that is sensitive to the relative magnitudes of $J_1^{\pi_\theta,\mu_\phi}$ and $J_2^{\mu_\phi}$ during training and is difficult to implement in a numerically stable manner.

To address this issue, we take the logarithm of the ratio. This transformation requires both terms to be strictly positive, so we introduce the following assumption.
\begin{assumption}
\label{assum:positivity}
For every admissible pair of policies $(\pi_\theta,\mu_\phi)$, $J_1^{\pi_\theta,\mu_\phi}>0$ and $J_2^{\mu_\phi}>0$.
\end{assumption}

Since $J_2^{\mu_\phi}$ is a cumulative squared norm, it is nonnegative by construction, and it is strictly positive whenever the adversary applies a nonzero disturbance with positive probability. Likewise, $J_1^{\pi_\theta,\mu_\phi}$ is strictly positive whenever the stage cost is nonnegative and is not almost surely zero along the trajectories. The quadratic costs used in our experiments (Section~\ref{sec:experiments}) are nonnegative, and the random initial state makes the accumulated cost nonzero in general.

Under Assumption~\ref{assum:positivity}, we define the transformed objective
\begin{align*}
    L(\theta,\phi) := \ln J^{\pi_\theta,\mu_\phi}
    = \ln J_1^{\pi_\theta,\mu_\phi} - \ln J_2^{\mu_\phi}.
\end{align*}
Since the logarithm is strictly increasing, this transformation leaves each player's best response unchanged: for a fixed $\mu_\phi$, the minimizers over $\theta$ are preserved, and for a fixed $\pi_\theta$, the maximizers over $\phi$ are preserved. It also converts the ratio into a difference of two terms, which simplifies the gradient-based optimization. Since $J_2^{\mu_\phi}$ depends on $\theta$ only through the state distribution, we neglect this dependence, and the gradients of $L$ with respect to $\theta$ and $\phi$ are given by
\begin{align}
    \label{eq:L1}
    \nabla_\theta  L(\theta,\phi)
    &= \frac{\nabla_\theta J_1^{\pi_\theta,\mu_\phi}}{J_1^{\pi_\theta,\mu_\phi}}, \\
    \nabla_\phi L(\theta,\phi)
    &= \frac{\nabla_\phi J_1^{\pi_\theta,\mu_\phi}}{J_1^{\pi_\theta,\mu_\phi}} - \frac{\nabla_\phi J_2^{\mu_\phi}}{J_2^{\mu_\phi}}.  \label{eq:L2}
\end{align}

\begin{remark}
\label{rem:mechanism}
Equation~\eqref{eq:L1} shows that $\nabla_\theta L(\theta,\phi)$ is a positive scalar multiple of $\nabla_\theta J_1^{\pi_\theta,\mu_\phi}$ and does not involve $J_2^{\mu_\phi}$. Hence, for a fixed adversary, the controller's descent direction coincides with that of the unregularized objective $J_1^{\pi_\theta,\mu_\phi}$, and only the step size is rescaled by $1/J_1^{\pi_\theta,\mu_\phi}$. The ratio objective therefore acts through the adversary. In~\eqref{eq:L2}, the relative gain in cost, $\nabla_\phi J_1^{\pi_\theta,\mu_\phi}/J_1^{\pi_\theta,\mu_\phi}$, is offset by the relative growth in the squared disturbance norm, $\nabla_\phi J_2^{\mu_\phi}/J_2^{\mu_\phi}$, so that the adversary locally benefits from a larger disturbance only when the cost grows proportionally faster than the squared disturbance norm. As a result, this adversary has a weaker incentive to enlarge the disturbance than an unregularized adversary that maximizes $J_1^{\pi_\theta,\mu_\phi}$ alone, and the controller is shaped indirectly through the disturbances it encounters during training.
\end{remark}

Furthermore, the gradients of $J_1^{\pi_\theta,\mu_\phi}$ and $J_2^{\mu_\phi}$ can be expressed using the deterministic policy gradient theorem~\cite{DPG} as
\begin{align*}
    \nabla_\theta J_1^{\pi_\theta,\mu_\phi}
    &= \mathbb{E}_{s \sim \rho^{\pi_\theta,\mu_\phi}} \Big[ \nabla_\theta \pi_\theta(s) \\
    &\qquad \times \nabla_a Q^{\pi_\theta,\mu_\phi}(s,a,\mu_\phi(s)) \big|_{a=\pi_\theta(s)} \Big], \\
    \nabla_\phi J_1^{\pi_\theta,\mu_\phi}
    &= \mathbb{E}_{s \sim \rho^{\pi_\theta,\mu_\phi}} \Big[ \nabla_\phi \mu_\phi(s) \\
    &\qquad \times \nabla_w Q^{\pi_\theta,\mu_\phi}(s,\pi_\theta(s),w) \big|_{w=\mu_\phi(s)} \Big], \\
    \nabla_\phi J_2^{\mu_\phi}
    &= \mathbb{E}_{s \sim \rho^{\pi_\theta,\mu_\phi}} \Big[ \nabla_\phi \mu_\phi(s) \\
    &\qquad \times \nabla_w Q^{\mu_\phi}(s,w) \big|_{w=\mu_\phi(s)} \Big],
\end{align*}
where $\rho^{\pi_\theta,\mu_\phi}$ is the discounted state visitation distribution induced by $\pi_\theta,\mu_\phi$, and $Q^{\pi_\theta,\mu_\phi}$, $Q^{\mu_\phi}$ are the action-value functions
\begin{multline*}
Q^{\pi_\theta,\mu_\phi}(s, a, w) \\= \mathbb{E} \left[ \left. \sum_{k=0}^{\infty} \gamma^{k} c_{k+1} \right| s_0 = s, a_0 = a, w_0=w, \pi_\theta, \mu_\phi \right],
\end{multline*}
\begin{multline*}
 Q^{\mu_\phi}(s, w) = \mathbb{E} \left[ \left. \sum_{k=0}^{\infty} \gamma^{k} \|w_k\|_2^2 \right|  s_0 = s, w_0 = w,\mu_\phi \right].
\end{multline*}

Using~\eqref{eq:L1} and~\eqref{eq:L2}, the controller updates $\theta$ via gradient descent, while the adversary updates $\phi$ via gradient ascent:
\begin{align*}
\theta_{k+1}&=\theta_k - \alpha_{\mathrm{con}} \left. \nabla_\theta L(\theta,\phi)\right|_{\theta=\theta_k}, \\
\phi_{k+1} &= \phi_k + \alpha_{\mathrm{adv}} \left. \nabla_\phi L(\theta,\phi)\right|_{\phi=\phi_k},
\end{align*}
where $\alpha_{\mathrm{con}},\alpha_{\mathrm{adv}}>0$ are step sizes.

Under deterministic policies $\pi_\theta,\mu_\phi$, the action-value functions satisfy the Bellman equations
\begin{align}
Q^{\pi_\theta,\mu_\phi}(s,a,w) &= \mathbb{E}\!\left[ c(s,a,w,s') \right. \nonumber\\
 & \left. + \gamma\, Q^{\pi_\theta,\mu_\phi} \bigl(s',\pi_\theta(s'),\mu_\phi(s')\bigr) \right], \label{eq:Q1}\\
Q^{\mu_\phi}(s,w) &= \mathbb{E}\!\left[ \|w\|_2^2 + \gamma\, Q^{\mu_\phi}  \bigl(s',\mu_\phi(s')\bigr) \right]. \label{eq:Q2}
\end{align}
These Bellman equations lead to a SARSA-type temporal-difference (TD) learning rule under deterministic policies. In the next section, we instantiate this scheme with neural network function approximators for $\pi_\theta$, $\mu_\phi$, $Q^{\pi_\theta,\mu_\phi}$, and $Q^{\mu_\phi}$, and describe how the resulting mini-batch estimates are used in place of the population-level quantities above.

\section{Implementation with Deep Deterministic Policy Gradient}
We implement these actor and critic updates using the network architecture and training techniques of DDPG~\cite{DDPG}. The resulting DDPG-based instantiation of CoDRA is presented below.

\subsection{Actor update}
Following the DDPG framework~\cite{DDPG}, we employ neural network critics $Q_{\psi_1}$ and $Q_{\psi_2}$, parameterized by $\psi_1$ and $\psi_2$, to approximate the action-value functions $Q^{\pi_\theta,\mu_\phi}$ and $Q^{\mu_\phi}$, respectively.
Consequently, the objectives $J_1^{\pi_\theta,\mu_\phi}$ and $J_2^{\mu_\phi}$ can be expressed as expectations of the corresponding action-value functions:
\begin{align*}
    J_1^{\pi_\theta,\mu_\phi}
    &= \mathbb{E}_{s_0 \sim \rho_0}
    \bigl[ Q^{\pi_\theta,\mu_\phi}(s_0,\pi_\theta(s_0),\mu_\phi(s_0)) \bigr],\\
    J_2^{\mu_\phi}
    &= \mathbb{E}_{s_0 \sim \rho_0}
    \bigl[ Q^{\mu_\phi}(s_0,\mu_\phi(s_0)) \bigr],
\end{align*}
where $\rho_0$ denotes the initial state distribution.

In practice, these expectations are approximated using mini-batch samples drawn from the replay buffer $\mathcal{D}$. Given a mini-batch $B=\{(s,a,w,s')\}$ sampled uniformly from $\mathcal{D}$, we estimate the expectations of the action-value functions using the current policies $\pi_\theta,\mu_\phi$ evaluated at the sampled states, and use the resulting batch means as scale estimates of $J_1^{\pi_\theta,\mu_\phi}$ and $J_2^{\mu_\phi}$:
\begin{align*}
 M(Q_{\psi_1}) &= \frac{1}{|B|} \sum_{s\in B} Q_{\psi_1}\bigl(s,\pi_\theta(s),\mu_\phi(s)\bigr), \\
 M(Q_{\psi_2}) &= \frac{1}{|B|} \sum_{s\in B} Q_{\psi_2}\bigl(s,\mu_\phi(s)\bigr),
\end{align*}
where $|B|$ denotes the mini-batch size. Note that $M(Q_{\psi_1})$ and $M(Q_{\psi_2})$ are themselves functions of $\theta$ and $\phi$ through $\pi_\theta(s)$ and $\mu_\phi(s)$. Whenever they appear as normalizing denominators below, we treat them as constants via a stop-gradient operator $\mathrm{sg}[\cdot]$, i.e., gradients are backpropagated through the numerator terms only.
Substituting these estimates into the logarithmic gradients in (\ref{eq:L1}) and (\ref{eq:L2}) yields the following mini-batch approximations:
\begin{align}
    \label{eq:Lappox}
    &\nabla_\theta  L(\theta,\phi)  \approx \frac{1}{|B|} \sum_{s\in B} \frac{\nabla_\theta Q_{\psi_1}\bigl(s,\pi_\theta(s),\mu_\phi(s)\bigr)}{\bigl|\mathrm{sg}[M(Q_{\psi_1})]\bigr|+\epsilon}, \\ \nonumber
    &\nabla_\phi  L(\theta,\phi)  \approx  \frac{1}{|B|} \sum_{s\in B} \left( \frac{\nabla_\phi Q_{\psi_1}(s,\pi_\theta(s),\mu_\phi(s)) }{\bigl|\mathrm{sg}[M(Q_{\psi_1})]\bigr|+\epsilon} \right. \\ & \qquad\qquad\qquad\qquad\qquad \left.- \frac{\nabla_\phi Q_{\psi_2}(s,\mu_\phi(s)) }{\bigl|\mathrm{sg}[M(Q_{\psi_2})]\bigr|+\epsilon} \right) .
    \label{eq:Lappox2}
\end{align}
Since the critics are randomly initialized, $M(Q_{\psi_1})$ and $M(Q_{\psi_2})$ can be negative early in training; the absolute value keeps each denominator positive, so the normalization never flips the sign of a gradient. The offset, fixed to $\epsilon=1$ in all experiments, keeps the denominator bounded away from zero when the batch-mean value approaches zero. This normalization is therefore intended to avoid exploding gradients and unintended sign flips, and to keep the competition between the controller and the adversary balanced when Q-values fluctuate during training.
To implement these gradient approximations in practice, we define the following joint actor loss $L(\theta, \phi; B)$:
\begin{multline*}
    L(\theta,\phi;B):=  \\
    \frac{1}{|B|} \sum_{s\in B} \left[\frac{Q_{{\psi_1}}(s,\pi_\theta(s),\mu_\phi(s))}{\bigl|\mathrm{sg}[M(Q_{\psi_1})]\bigr|+\epsilon} - \frac{Q_{{\psi_2}}(s,\mu_\phi(s))}{\bigl|\mathrm{sg}[M(Q_{\psi_2})]\bigr|+\epsilon} \right].
\end{multline*}
The online actor parameters $\theta$ and $\phi$ are updated using these sampled deterministic policy gradients~\cite{DDPG,DPG}:
\begin{align*}
\theta &\leftarrow \theta - \alpha_{\mathrm{con}} \nabla_\theta L(\theta,\phi;B),\\
\phi   &\leftarrow \phi + \alpha_{\mathrm{adv}} \nabla_\phi L(\theta,\phi;B).
\end{align*}
After updating the online actors, the target parameters $\theta'$ and $\phi'$ are softly updated as
\begin{align*}
\theta' &\leftarrow \tau \theta + (1-\tau)\theta', \\
\phi'   &\leftarrow \tau \phi + (1-\tau)\phi',
\end{align*}
where $\tau \in (0,1)$ is the interpolation coefficient controlling the update rate. As in DDPG, this soft update keeps the target policies slowly varying.

\begin{algorithm}[!h]
\caption{Cost-to-disturbance ratio adversarial training (CoDRA) with DDPG}\label{alg:CoDRA}
\begin{algorithmic}[1]
\State Initialize the online critic networks $Q_{\psi_1},Q_{\psi_2}$
\State Initialize the online actor networks $\pi_{\theta},  \mu_{\phi}$ for the controller and adversary, respectively.
\State Initialize the target parameters $\psi_1'\gets\psi_1, \psi_2' \gets \psi_2$, $\quad\theta' \gets \theta , \phi' \gets \phi$
\State Initialize the replay buffer $\mathcal{D}$

\For{ Episode $i=1,2,...N_{episode}$}
\State Sample the initial state $s_0 \sim \rho_0$
\For{ Time step $k=0,1,2,...T-1$}
\State The controller and adversary select an action and a disturbance
\State $a_k = \pi_{\theta}(s_k)+\xi_k^a, \quad w_k = \mu_{\phi}(s_k)+\xi_k^w$
\State where $\xi_k^a, \xi_k^w$ are Ornstein--Uhlenbeck exploration noise~\cite{OUnoise}, as in the standard DDPG approach~\cite{DDPG}.
\State Observe next state $s_{k+1}$
\State Compute the cost $c_{k+1}:=c(s_k,a_k,w_k,s_{k+1})$
\State Store $(s_k,a_k,w_k,c_{k+1},s_{k+1})$ in $\mathcal{D}$
\State Uniformly sample a mini-batch $B$ from $\mathcal{D}$
\State Update the online critic parameters according to (\ref{eq:Lpsi1}) and (\ref{eq:Lpsi2}):
\begin{align*}
    {\psi _1} &\gets {\psi _1} - {\alpha _{{\rm{critic}}}}{\nabla _{\psi_1} }L({\psi_1};B)\\
    {\psi _2} &\gets {\psi _2} - {\alpha _{{\rm{critic}}}}{\nabla _{\psi_2} }L({\psi_2};B)
\end{align*}
\State Update the online actor parameters according to (\ref{eq:Lappox}) and (\ref{eq:Lappox2}):
\begin{align*}
\theta  &\gets \theta  - {\alpha _{\rm con}}{\nabla _{\theta} }L(\theta ,\phi ;B),\\
\phi  &\gets \phi  + {\alpha_{\rm adv}}{\nabla _{\phi} }L(\theta ,\phi ;B)
\end{align*}

\State Soft update target networks:
\[
\theta' \gets \tau\theta +(1-\tau)\theta', \quad
\phi' \gets \tau\phi +(1-\tau)\phi'
\]
\[
\psi_1' \gets \tau \psi_1 + (1-\tau)\psi_1', \quad
\psi_2' \gets \tau \psi_2 + (1-\tau)\psi_2'
\]
\EndFor
\EndFor
\end{algorithmic}
\end{algorithm}

\subsection{Critic update}
Following DDPG~\cite{DDPG}, the critics $Q_{\psi_1}$ and $Q_{\psi_2}$ are trained by approximating the Bellman equations (\ref{eq:Q1}) and (\ref{eq:Q2}) with sampled transitions, which yields a mini-batch SARSA-style TD learning rule.
Given a mini-batch of transitions $B=\{(s,a,w,s')\}$ sampled from $\mathcal{D}$, the critic losses are defined as the mean squared TD errors
\begin{align*}
L(\psi_1;B) &=\frac{1}{2}\frac{1}{|B|}  \sum_{(s,a,w,s')\in B}  \left( y_1 - Q_{\psi_1}(s,a,w)\right)^2,\\
L(\psi_2;B) &= \frac{1}{2} \frac{1}{|B|}\sum_{(s,a,w,s')\in B} \left( y_2 - Q_{\psi_2}(s,w) \right)^2,
\end{align*}
where the targets are defined as follows:
\begin{align*}
y_1 &=c(s,a,w,s')+\gamma\,\mathbf{1}(s')\, Q_{\psi_1'}\!\left(s',\pi_{\theta'}(s'),\mu_{\phi'}(s')\right),\\
y_2 &=\|w\|_2^2 +\gamma\,\mathbf{1}(s')\, Q_{\psi_2'}\!\left(s',\mu_{\phi'}(s')\right),
\end{align*}
where $\mathbf{1}(s')=0$ if $s'$ is a terminal state and $\mathbf{1}(s')=1$ otherwise.
The online critic parameters are updated via gradient descent:
\begin{align*}
\psi_1 &\leftarrow \psi_1 - \alpha_{\mathrm{critic}}\nabla_{\psi_1} L(\psi_1;B),\\
\psi_2 &\leftarrow \psi_2 - \alpha_{\mathrm{critic}}\nabla_{\psi_2} L(\psi_2;B),
\end{align*}
where $\alpha_{\mathrm{critic}}>0$ is the learning rate, and the gradients of the critic losses are given by
\begin{align}
    \label{eq:Lpsi1}
    &\nabla_{\psi_1} L(\psi_1;B) = \nonumber \\ &\frac{1}{|B|} \sum_{(s,a,w,s')\in B} \bigl( Q_{{\psi_1}}(s,a,w) - y_1 \bigr) \nabla_{\psi_1} Q_{\psi_1} \bigl(s,a,w\bigr),
\end{align}
\begin{align}
    &\nabla_{\psi_2} L(\psi_2;B) = \nonumber \\ &\frac{1}{|B|} \sum_{(s,a,w,s')\in B} \bigl( Q_{{\psi_2}}(s,w) - y_2 \bigr) \nabla_{\psi_2} Q_{\psi_2} \bigl(s,w\bigr) .
    \label{eq:Lpsi2}
\end{align}
After updating the online critics, the target critic parameters are softly updated as
\begin{align*}
\psi_1' &\leftarrow \tau \psi_1 + (1-\tau)\psi_1',\\
\psi_2' &\leftarrow \tau \psi_2 + (1-\tau)\psi_2',
\end{align*}
where $\tau\in(0,1)$ is the interpolation coefficient.

The overall algorithm is described in Algorithm~\ref{alg:CoDRA}.

\section{Experiment and results}
\label{sec:experiments}

\subsection{Environments}
\label{sec:environments}
We evaluate the proposed algorithm on two MuJoCo cart-pole stabilization tasks~\cite{Mujoco} of increasing difficulty: \texttt{InvertedPendulum-v5} (single pendulum) and \texttt{InvertedDoublePendulum-v5} (two-link pendulum). Both use a nonnegative quadratic cost in place of the original Gymnasium reward, which supports the positivity of $J_1^{\pi_\theta,\mu_\phi}$ in Assumption~\ref{assum:positivity}.

InvertedPendulum uses stage cost $c(s,a) = \theta^2 + 0.1\, x^2 + 0.001\, a^2$, where $x$ is the cart position and $\theta$ is the pole angle from vertical, over fixed 100-step episodes with no early termination. InvertedDoublePendulum uses $c(s,a) = \theta_1^2 + \theta_2^2 + 0.1\, x^2 + 0.001\, a^2$, where $\theta_1,\theta_2$ are the angles of the lower and upper (\textit{pole}, \textit{pole2}) links, and additionally terminates the episode early with a one-time penalty of 10.0 whenever the tip height falls to or below $z_{\text{tip}} = 1.0\,\mathrm{m}$, over a maximum of 200 steps.

In both environments, the adversarial methods (CoDRA, RARL) act through a scalar lateral force applied directly to the pole body (InvertedPendulum, $w \in [-2.5, 2.5]$) or to \textit{pole2} (InvertedDoublePendulum, $w \in [-5, 5]$) at every simulation step, with the same bound for both methods, independent of the control actuator, while domain-randomization baselines instead rescale the link masses once per episode.

In both tasks the state is the concatenation of the MuJoCo joint positions and velocities: $s=(x,\theta,\dot x,\dot\theta)\in\mathbb{R}^4$ for InvertedPendulum and $s=(x,\theta_1,\theta_2,\dot x,\dot\theta_1,\dot\theta_2)\in\mathbb{R}^6$ for InvertedDoublePendulum. The action is a scalar command to the cart motor, and each policy action is held for two simulation steps. Every episode starts from $s_0\sim\mathcal{U}(-0.01,0.01)$ componentwise. Table~\ref{tab:sim} lists the remaining simulation parameters.

\begin{table}[t]
    \centering
    \caption{Simulation parameters.}
    \label{tab:sim}
    \begin{tabular}{lcc}
        \toprule
        & InvertedPendulum & InvertedDoublePendulum \\
        \midrule
        State dimension & $4$ & $6$ \\
        Action range & $[-3,3]$ & $[-1,1]$ \\
        Motor gear & $100$ & $500$ \\
        Integrator, timestep & RK4, $0.02$\,s & RK4, $0.01$\,s \\
        Frame skip & $2$ & $2$ \\
        Control period & $0.04$\,s & $0.02$\,s \\
        Episode length & $100$ steps & $\le 200$ steps \\
        Early termination & none & $z_{\text{tip}}\le 1.0$\,m ($+10$) \\
        Cart rail range & $[-1,1]$\,m & $[-1,1]$\,m \\
        Hinge range & $\pm 90^\circ$ & unlimited \\
        Joint damping & $1$ & $0.05$ \\
        Link length & $0.6$\,m & $0.6$\,m each \\
        Disturbance body & pole & \textit{pole2} \\
        Adversary bound & $\pm 2.5$ & $\pm 5$ \\
        \bottomrule
    \end{tabular}
\end{table}

\subsection{Experiment setup}
On both environments, we compare CoDRA against six baselines that span the two prior-art families discussed in Section~\ref{sec:relatedworks}. The main comparison uses five of them; the random-force ablation DDPG-RF is reported separately in Section~\ref{sec:results-all}.
\begin{itemize}
\item \textbf{DDPG}~\cite{DDPG}: trained on the nominal plant without any perturbation.
\item \textbf{DR-DDPG}: DDPG whose link masses are rescaled at the start of every episode by factors drawn log-uniformly from $[0.25,8]$ (pole) and $[0.5,2]$ (cart) on InvertedPendulum, and from $[0.5,2]$ (both links) on InvertedDoublePendulum. These ranges cover the evaluation mass grids below.
\item \textbf{RARL}~\cite{RARL}: a learned adversary that uses the same force channel and bound as CoDRA, but maximizes the cumulative cost without normalization. On both environments, the two policies are updated in alternating blocks of $10$ episodes, following~\cite{RARL}: while one policy is updated, the other is held fixed.
\item \textbf{PR-DDPG} and \textbf{NR-DDPG}~\cite{ARDDPG}: action-robust baselines whose adversary acts through the action channel rather than through an external force. In PR-DDPG the adversary's action replaces the controller's with probability $\alpha$; in NR-DDPG the executed action is $(1-\alpha)a+\alpha w$. We use $\alpha=0.1$ on both environments.
\item \textbf{DDPG-RF}: a random-force ablation. DDPG is trained while a non-learned lateral force drawn from $\mathcal{N}(0,3^2)$ is applied at every step to the same body the adversary acts on. It tests whether any persistent random force suffices, or whether the force must be learned.
\end{itemize}
All algorithms share the same network architecture and optimization settings (Table~\ref{tab:hyper}) and are trained over twenty independent random seeds; every policy is evaluated at its final checkpoint. All learned adversaries are removed at evaluation: the force sweep instead applies a constant, non-learned force, and the mass grid applies no disturbance at all, so both axes measure transfer rather than rejection of the matched adversarial disturbance the controller was trained against.

For each environment we evaluate robustness along two axes: a force sweep and a mass grid, both measured by the mean discounted cost; for InvertedDoublePendulum, we also report the fall rate, i.e., the fraction of evaluation episodes that terminate early.

\textbf{Force sweep:} at evaluation time we apply a constant (non-adversarial, non-learned) lateral force to the pole (InvertedPendulum) or \textit{pole2} (InvertedDoublePendulum) at the nominal mass. The magnitude $F$ is swept over $\{0,0.2,0.5,1,2,4,6\}$. The largest magnitudes exceed the bound imposed on the adversary during training: $4$ and $6$ for InvertedPendulum ($\pm 2.5$), and $6$ for InvertedDoublePendulum ($\pm 5$). Since this force is fixed rather than optimized against the controller, the sweep probes a held-out disturbance and is not a worst-case evaluation.

\textbf{Mass grid:} we perturb the link masses of the plant at evaluation time, with no disturbance applied. For InvertedPendulum, we scale the pole mass by factors in $\{0.25,0.5,1.0,2.0,4.0\}$ and the cart mass by factors in $\{0.5,1.0,2.0\}$; for most methods, the lightest pole ($0.25\times$) is the most challenging setting. For InvertedDoublePendulum we apply the analogous $\{0.5,1.0,2.0\}\times\{0.5,1.0,2.0\}$ scaling separately to the lower-link (\textit{pole}) and upper-link (\textit{pole2}) masses. Such mass uncertainty is representative of payload variation and imperfect system identification in real deployments. Each cell of the force sweep and the mass grid is averaged over $100$ evaluation episodes per seed on InvertedPendulum and $50$ on InvertedDoublePendulum, and the reported value is the mean over seeds.

\begin{table}[t]
    \centering
    \caption{Training hyperparameters, shared by all algorithms unless noted.}
    \label{tab:hyper}
    \begin{tabular}{ll}
        \toprule
        Hidden layers & $2\times 256$, ReLU \\
        Output layer & $\tanh$, scaled to the action/force bound \\
        Optimizer & Adam \\
        Actor / adversary learning rate & $10^{-4}$ \\
        Critic learning rate & $10^{-3}$ \\
        Discount factor $\gamma$ & $0.99$ \\
        Mini-batch size $|B|$ & $128$ \\
        Replay buffer size & $10^6$ \\
        Target update rate $\tau$ & $0.005$ \\
        Exploration noise & OU, $\theta=0.15$, $\sigma=0.2$ \\
        Training steps & $10^5$ \\
        Updates begin after & $2|B|$ transitions \\
        Checkpoint interval & $2\times10^4$ steps (final one evaluated) \\
        Seeds & $20$ \\
        \midrule
        \multicolumn{2}{l}{CoDRA only} \\
        Normalization offset $\epsilon$ & $1$ \\
        Controller : adversary updates per step & $2:1$ \\
        \bottomrule
    \end{tabular}
\end{table}

\subsection{Results}
\subsubsection{InvertedPendulum}
\label{sec:results-ip}

\begin{table}[t]
    \centering
    \caption{InvertedPendulum: mean discounted cost at the nominal plant, under the largest force, and in the worst mass cell. \emph{Nominal}: force-$0$ entry of the force sweep (unit mass scale). \emph{Force $=6$}: fixed lateral pole force, nominal mass. \emph{Worst mass cell}: max over the pole $\{0.25,0.5,1,2,4\}\times$ cart $\{0.5,1,2\}$ grid. Lower is better; bold is best per column.}
    \label{tab:ip-summary}
    \begin{tabular}{lccc}
        \toprule
        Algorithm & Nominal & Force $=6$ & Worst mass cell \\
        \midrule
        DDPG~\cite{DDPG}      & \cellcolor{blue!0}0.61  & \cellcolor{blue!1}28.76 & \cellcolor{blue!31}19.40 \\
        DR-DDPG                & \cellcolor{blue!2}0.78 & \cellcolor{blue!70}53.59 & \cellcolor{blue!0}\textbf{11.78} \\
        RARL~\cite{RARL}      & \cellcolor{blue!0}\textbf{0.59}  & \cellcolor{blue!0}\textbf{28.48} & \cellcolor{blue!42}21.85 \\
        PR-DDPG~\cite{ARDDPG} & \cellcolor{blue!34}4.24 & \cellcolor{blue!41}43.33 & \cellcolor{blue!35}20.28 \\
        NR-DDPG~\cite{ARDDPG} & \cellcolor{blue!70}8.17 & \cellcolor{blue!17}34.66 & \cellcolor{blue!70}28.73 \\
        CoDRA (ours)         & \cellcolor{blue!12}1.91 & \cellcolor{blue!16}34.14 & \cellcolor{blue!20}16.67 \\
        \bottomrule
    \end{tabular}
\end{table}

Table~\ref{tab:ip-summary} summarizes these three quantities. Unlike InvertedDoublePendulum below, CoDRA is not the best method here. Its nominal cost ($1.91$) is higher than those of DDPG, DR-DDPG, and RARL, although well below those of the two action-robust baselines ($4.24$ and $8.17$). Under the largest force ($F=6$) it is third of six, behind RARL and DDPG, and on the mass grid it is second, behind DR-DDPG.

This suggests CoDRA's advantage is specific to the harder InvertedDoublePendulum task with early termination, not a general property of the method: on the milder InvertedPendulum, where DDPG already recovers well from perturbation and a fall does not end the episode, plain domain randomization or a well-tuned unregulated adversary (RARL) are at least as effective.

\subsubsection{InvertedDoublePendulum}
\label{sec:results-dp}

\begin{figure}[t]
    \centering
    \includegraphics[width=\linewidth]{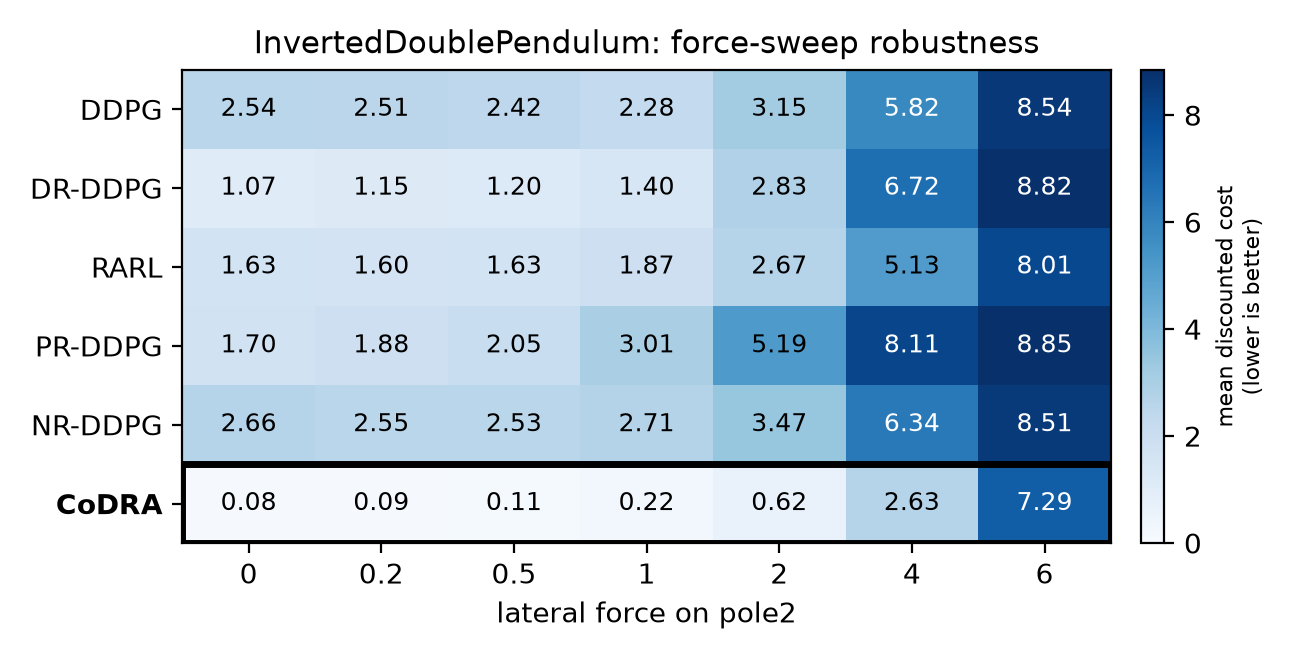}
    \caption{InvertedDoublePendulum: mean discounted cost under a constant lateral force on \textit{pole2}, swept from $0$ to $6$ at the nominal mass. The adversary was restricted to $\pm 5$ during training, so the largest force lies outside that range. CoDRA (bold border) has the lowest cost at every force level; the gap to the baselines is largest up to $F=4$ and narrows at $F=6$.}
    \label{fig:dp-force}
\end{figure}

\begin{figure}[t]
    \centering
    \includegraphics[width=\linewidth]{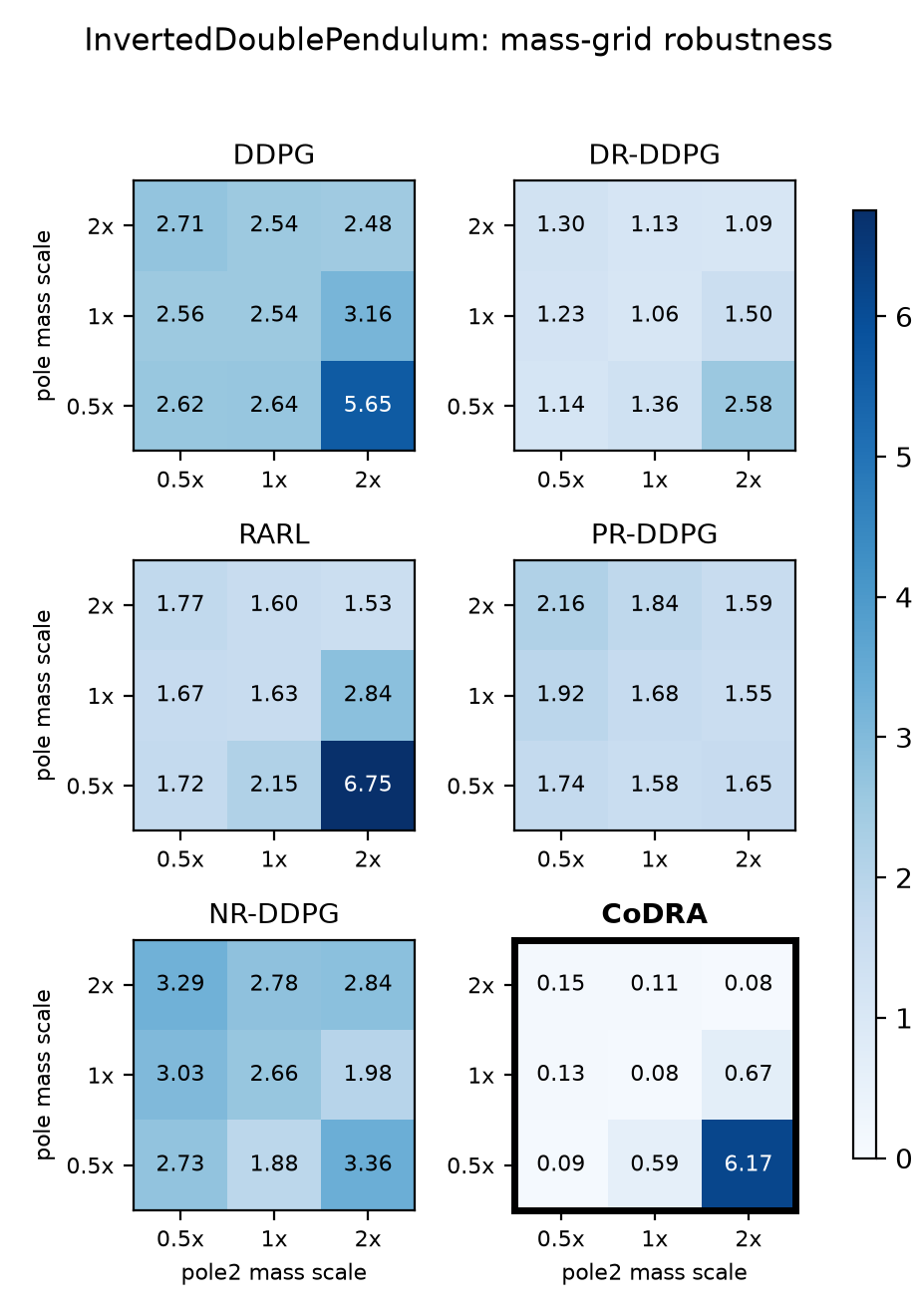}
    \caption{InvertedDoublePendulum: mean discounted cost over the pole $\times$ pole2 mass $\{0.5,1,2\}\times\{0.5,1,2\}$ grid, nominal at $1\times/1\times$, with no disturbance applied. Darker is higher (worse) cost. CoDRA (bold border) is lowest in every cell except (pole $0.5\times$, pole2 $2\times$), where PR-DDPG is lowest.}
    \label{fig:dp-mass}
\end{figure}

Figures~\ref{fig:dp-force} and~\ref{fig:dp-mass} tell a different story from InvertedPendulum: CoDRA has by far the lowest nominal cost ($0.08$, about $13\times$ below the best baseline, DR-DDPG, and $30\times$ below DDPG) and the lowest cost at every force level. As the force increases to $F=4$, CoDRA remains well below every baseline ($2.63$ at $F=4$, compared with $5.13$ for the best baseline, RARL). The fall rate shows the same trend: CoDRA does not fall in any episode up to $F=2$ and falls in $8\%$ of the episodes at $F=4$, whereas the baselines fall in $54$--$100\%$ of the episodes at $F=4$. At $F=6$, which lies outside the $\pm 5$ training range, all methods degrade and the gap narrows, but CoDRA still has the lowest cost ($7.29$, compared with $8.01$ for RARL) and the lowest fall rate ($78\%$, compared with $87$--$100\%$).

The mass grid shows the same pattern in $8$ of $9$ cells (CoDRA stays below $0.7$ while the best of the other five is $2$--$14\times$ higher), except the (pole $0.5\times$, pole2 $2\times$) corner. There CoDRA's cost ($6.17$) is the second highest of the six, below only RARL's ($6.75$), and PR-DDPG, whose cost stays nearly flat across the grid, is lowest ($1.65$). This corner pairs the heaviest tip link with the lightest lower link, which suggests that CoDRA's advantage is least reliable when the mass is concentrated at the tip.

This asymmetry is consistent with Remark~\ref{rem:mechanism}: since CoDRA's robustness comes from regulating the adversary's incentive rather than from the controller directly trading off performance against disturbance, its benefit would be expected to scale with how destabilizing an unregulated adversary would otherwise be---small on InvertedPendulum, where the controller recovers easily on its own, and large on InvertedDoublePendulum, where an unregulated adversary can drive the controller into a catastrophic, hard-to-recover fall.

\subsubsection{Random-force ablation}
\label{sec:results-all}

\begin{figure*}[t]
    \centering
    \includegraphics[width=0.8\textwidth]{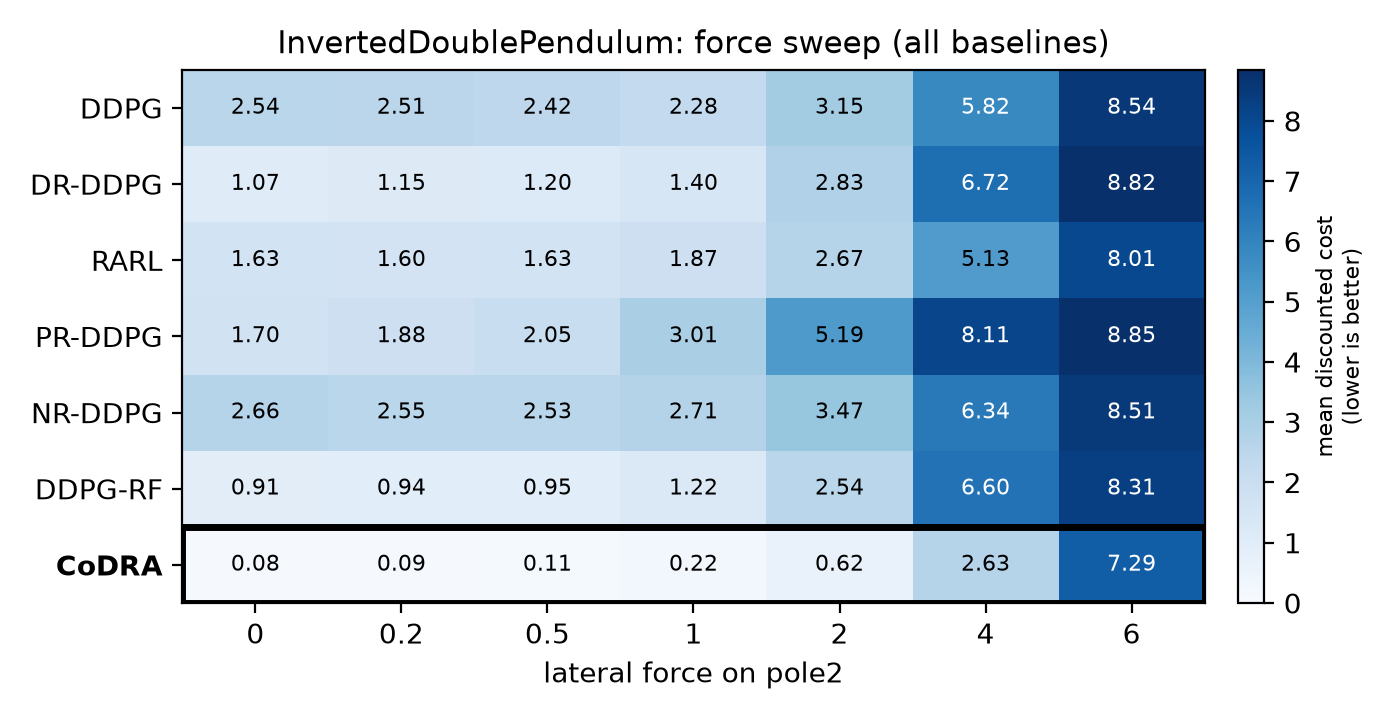}
    \caption{InvertedDoublePendulum force sweep with all seven methods. The adversarial methods were trained with a $\pm 5$ bound, so only $F=6$ lies outside the training range.}
    \label{fig:ext-dp-force}
\end{figure*}

\begin{figure*}[t]
    \centering
    \includegraphics[width=\textwidth]{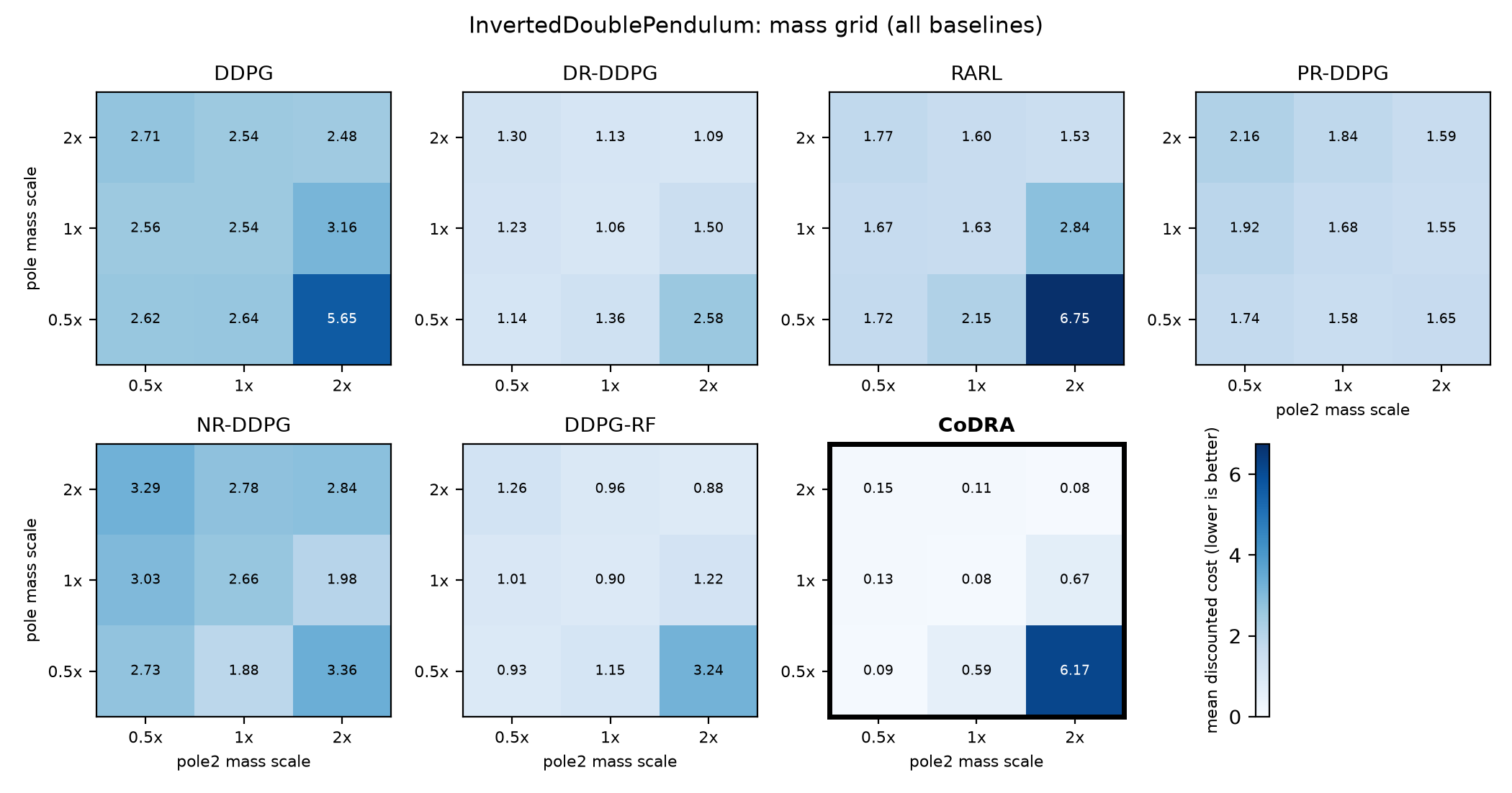}
    \caption{InvertedDoublePendulum mass grid with all seven methods, pole $\times$ pole2 mass $\{0.5,1,2\}\times\{0.5,1,2\}$, no disturbance applied.}
    \label{fig:ext-dp-mass}
\end{figure*}

\begin{figure*}[t]
    \centering
    \includegraphics[width=0.8\textwidth]{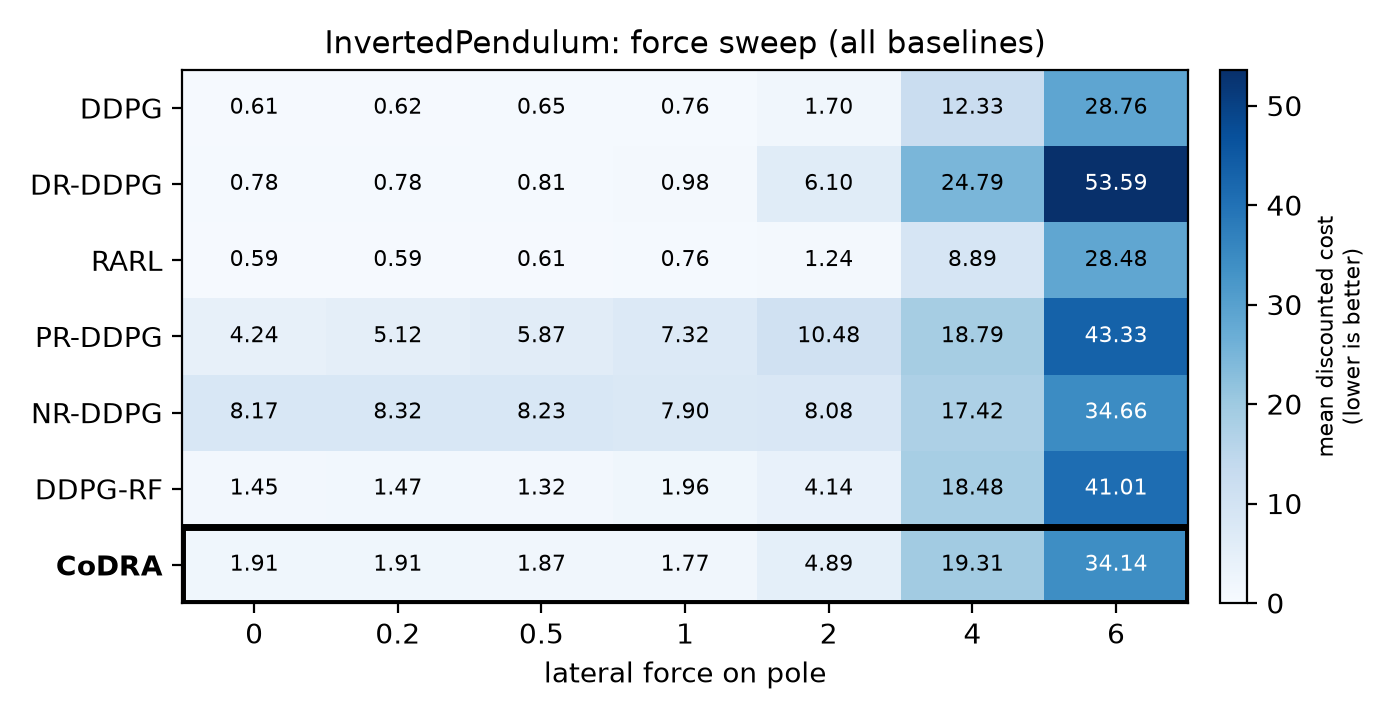}
    \caption{InvertedPendulum force sweep with all seven methods. The adversarial methods were trained with a $\pm 2.5$ bound, so $F=4$ and $F=6$ lie outside the training range.}
    \label{fig:ext-ip-force}
\end{figure*}

\begin{figure*}[t]
    \centering
    \includegraphics[width=\textwidth]{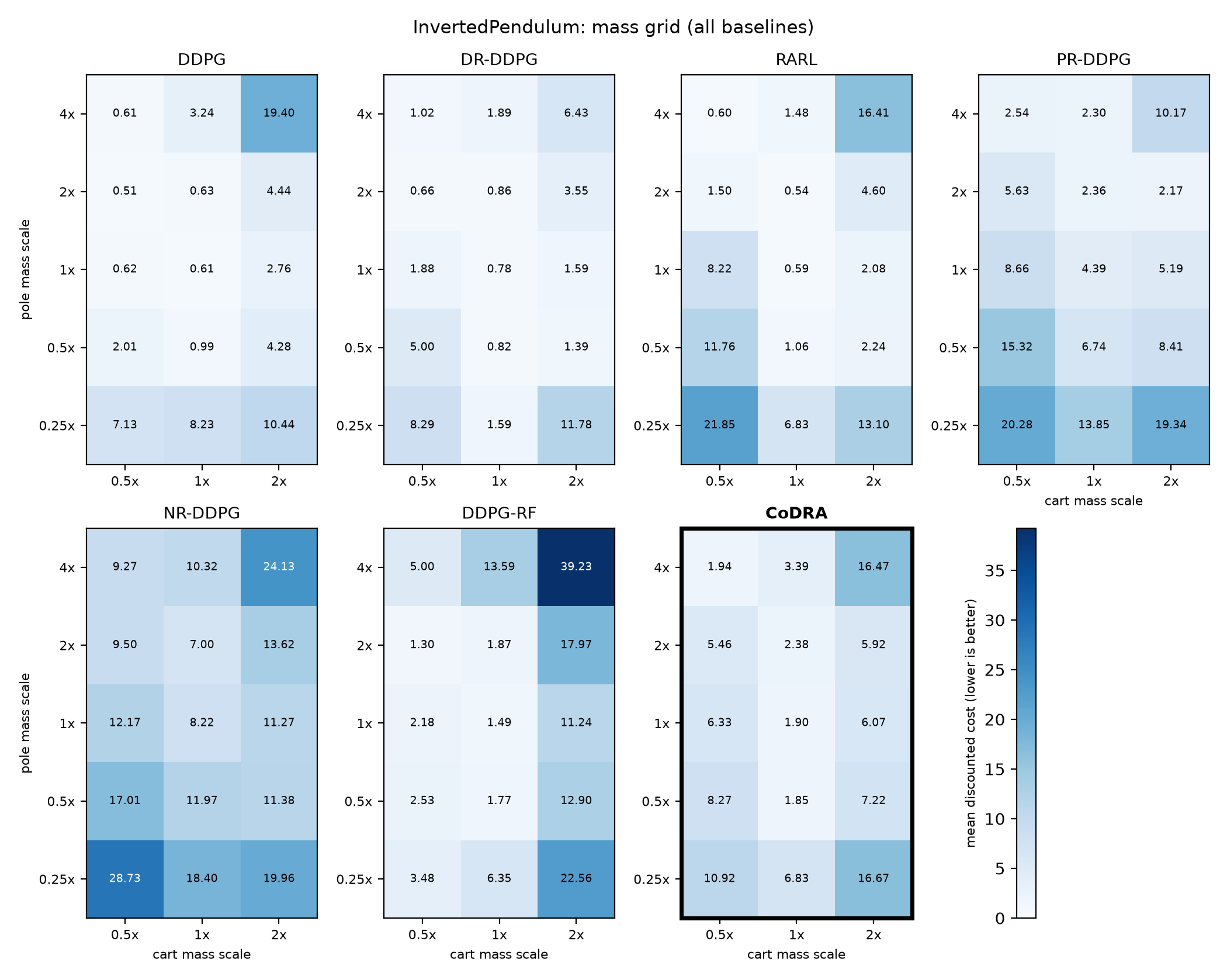}
    \caption{InvertedPendulum mass grid with all seven methods, pole $\{0.25,0.5,1,2,4\}\times$ cart $\{0.5,1,2\}$, no disturbance applied.}
    \label{fig:ext-ip-mass}
\end{figure*}

\begin{table}[t]
    \centering
    \caption{InvertedPendulum with all seven methods: mean discounted cost. \emph{Nominal} is the force-$0$ entry of the force sweep; the other columns are defined as in Table~\ref{tab:ip-summary}. Bold is best per column.}
    \label{tab:ip-all}
    \begin{tabular}{lccc}
        \toprule
        Algorithm & Nominal & Force $=6$ & Worst mass cell \\
        \midrule
        DDPG~\cite{DDPG}   & $0.61$ & $28.76$ & $19.40$ \\
        DR-DDPG            & $0.78$ & $53.59$ & $\mathbf{11.78}$ \\
        RARL~\cite{RARL}   & $\mathbf{0.59}$ & $\mathbf{28.48}$ & $21.85$ \\
        PR-DDPG~\cite{ARDDPG} & $4.24$ & $43.33$ & $20.28$ \\
        NR-DDPG~\cite{ARDDPG} & $8.17$ & $34.66$ & $28.73$ \\
        DDPG-RF            & $1.45$ & $41.01$ & $39.23$ \\
        CoDRA (ours)       & $1.91$ & $34.14$ & $16.67$ \\
        \bottomrule
    \end{tabular}
\end{table}

Figures~\ref{fig:ext-dp-force}--\ref{fig:ext-ip-mass} and Table~\ref{tab:ip-all} add the random-force ablation DDPG-RF to both evaluations. It asks whether CoDRA's robustness requires a learned, normalized adversary, or whether any persistent random force on the same body would suffice.

On InvertedDoublePendulum, DDPG-RF is the strongest of the six baselines at most conditions: it has the lowest baseline cost from $F=0$ to $F=2$ and in $8$ of the $9$ mass-grid cells. CoDRA is nevertheless lower at every force level (for example, $0.62$ versus $2.54$ at $F=2$) and in the same $8$ mass-grid cells. A persistent random force therefore improves substantially on plain DDPG, but it does not reproduce the effect of the learned, normalized adversary.

On InvertedPendulum (Table~\ref{tab:ip-all}), DDPG-RF has a lower nominal cost than CoDRA ($1.45$ versus $1.91$), but under the largest force ($41.01$) and on the mass grid ($39.23$, the largest worst-case cost of all methods) it is worse than CoDRA. Training under a strong random force thus does not carry over to mass mismatch on this task, and the ablation leaves the conclusion of Section~\ref{sec:results-ip} unchanged.

\subsubsection{Seed variability}
\label{sec:seed-stats}

\begin{table*}[t]
    \centering
    \caption{Mean $\pm$ standard deviation of the discounted cost over the $20$ seeds. $t$ is Welch's $t$-statistic between CoDRA and the baseline with the lowest mean (among all six baselines, including DDPG-RF); negative values favor CoDRA. The last column is the smallest $|t|$ over the six baselines. The InvertedPendulum mass cell is CoDRA's worst cell (pole $0.25\times$, cart $2\times$); grid means are computed per seed before averaging.}
    \label{tab:seed-stats}
    \begin{tabular}{llcccc}
        \toprule
        Environment & Quantity & CoDRA & Lowest-mean baseline & $t$ & $\min |t|$ \\
        \midrule
        InvertedDoublePendulum & $F=0$ & $0.08 \pm 0.09$ & DDPG-RF, $0.91 \pm 0.90$ & $-4.1$ & $3.7$ \\
         & $F=2$ & $0.62 \pm 0.11$ & DDPG-RF, $2.54 \pm 1.45$ & $-5.9$ & $5.1$ \\
         & $F=4$ & $2.63 \pm 0.74$ & RARL, $5.13 \pm 1.47$ & $-6.8$ & $6.8$ \\
         & $F=6$ & $7.29 \pm 1.20$ & RARL, $8.01 \pm 0.91$ & $-2.1$ & $2.1$ \\
         & Mass-grid mean & $0.89 \pm 0.71$ & DDPG-RF, $1.28 \pm 1.16$ & $-1.3$ & $1.3$ \\
         & Cell (pole $0.5\times$, pole2 $2\times$) & $6.17 \pm 4.07$ & PR-DDPG, $1.65 \pm 1.74$ & $+4.6$ & $0.4$ \\
        \midrule
        InvertedPendulum & $F=0$ & $1.91 \pm 3.92$ & RARL, $0.59 \pm 0.44$ & $+1.5$ & $0.5$ \\
         & $F=6$ & $34.14 \pm 22.84$ & RARL, $28.48 \pm 19.02$ & $+0.9$ & $0.1$ \\
         & Cell (pole $0.25\times$, cart $2\times$) & $16.67 \pm 20.49$ & DDPG, $10.44 \pm 15.94$ & $+1.1$ & $0.4$ \\
         & Mass-grid mean & $6.77 \pm 7.49$ & DR-DDPG, $3.17 \pm 3.33$ & $+2.0$ & $0.3$ \\
        \bottomrule
    \end{tabular}
\end{table*}

Table~\ref{tab:seed-stats} reports the seed-to-seed variability of the main quantities. With $20$ seeds per method, $|t|>2$ corresponds roughly to $p<0.05$ (two-sided, without correction for multiple comparisons). On InvertedDoublePendulum, CoDRA's lower cost up to $F=4$ is large relative to this variability, with $|t|\ge 3.7$ against every baseline. At $F=6$, the gap to RARL is marginal ($t=-2.1$). Averaged over the mass grid, CoDRA's lower mean cost is not statistically distinguishable from that of DDPG-RF ($t=-1.3$), and in the (pole $0.5\times$, pole2 $2\times$) corner CoDRA is significantly worse than PR-DDPG ($t=+4.6$). On InvertedPendulum, none of the differences between CoDRA and the lowest-mean baseline is significant at this level, although CoDRA's mass-grid mean is marginally higher than that of DR-DDPG ($t=+2.0$); this is consistent with the conclusion of Section~\ref{sec:results-ip} that CoDRA's advantage is less pronounced on this task.

\section{Conclusion}
This paper presented CoDRA, an adversarial training framework for learning continuous-control policies that remain robust to external disturbances. The proposed framework formulates policy learning as a two-player zero-sum game between a controller and an adversary, and expresses the training objective as a ratio of accumulated cost to accumulated squared disturbance norm. The adversary's disturbance magnitude is then regulated through a self-normalized actor-critic update, rather than through an explicit penalty term or an auxiliary trade-off parameter that must be tuned or adapted online.

Experiments on two MuJoCo cart-pole tasks with nonnegative quadratic costs show that CoDRA's benefit is not uniform. On InvertedDoublePendulum, where a fall terminates the episode with a penalty, CoDRA attains a lower cost than all compared baselines at every force level, including a force outside the range the adversary was allowed during training, and in all but one cell of the mass grid. On the milder InvertedPendulum, which has no early termination, its advantage is less pronounced. A random-force ablation shows that a persistent random force alone does not reproduce CoDRA's advantage on InvertedDoublePendulum. As discussed in Section~\ref{sec:results-dp}, this contrast is consistent with self-normalization regulating the adversary's incentive rather than the controller directly, which suggests that the benefit grows with how destabilizing an unregulated adversary would otherwise be.

Future work includes establishing formal attenuation guarantees for the ratio objective, understanding why the benefit concentrates on the task with early termination, broadening the evaluation to higher-dimensional robotic platforms, and extending the update to tasks whose cost can be negative, where Assumption~\ref{assum:positivity} no longer holds and the logarithmic transformation must be modified.

\bibliographystyle{IEEEtran}
\bibliography{ref}

@inproceedings{Mujoco,
  title={Mujoco: A physics engine for model-based control},
  author={Todorov, Emanuel and Erez, Tom and Tassa, Yuval},
  booktitle={2012 IEEE/RSJ international conference on intelligent robots and systems},
  pages={5026--5033},
  year={2012},
  organization={IEEE}
}

@inproceedings{perolat2015approximate,
  title={Approximate dynamic programming for two-player zero-sum Markov games},
  author={Perolat, Julien and Scherrer, Bruno and Piot, Bilal and Pietquin, Olivier},
  booktitle={International Conference on Machine Learning},
  pages={1321--1329},
  year={2015},
  organization={PMLR}
}

@article{RRL,
  title={Robust reinforcement learning},
  author={Morimoto, Jun and Doya, Kenji},
  journal={Neural computation},
  volume={17},
  number={2},
  pages={335--359},
  year={2005},
  publisher={MIT Press}
}

@inproceedings{DPG,
  title={Deterministic policy gradient algorithms},
  author={Silver, David and Lever, Guy and Heess, Nicolas and Degris, Thomas and Wierstra, Daan and Riedmiller, Martin},
  booktitle={International Conference on Machine Learning},
  pages={387--395},
  year={2014},
  organization={Pmlr}
}

@article{DDPG,
     title={Continuous control with deep reinforcement learning}, 
      author={Timothy P. Lillicrap and Jonathan J. Hunt and Alexander Pritzel and Nicolas Heess and Tom Erez and Yuval Tassa and David Silver and Daan Wierstra},
      journal={arXiv preprint arXiv:1509.02971},
      year={2019}
}

@article{RL1,
  title={Mastering the game of go without human knowledge},
  author={Silver, David and Schrittwieser, Julian and Simonyan, Karen and Antonoglou, Ioannis and Huang, Aja and Guez, Arthur and Hubert, Thomas and Baker, Lucas and Lai, Matthew and Bolton, Adrian and others},
  journal={Nature},
  volume={550},
  number={7676},
  pages={354--359},
  year={2017},
  publisher={Nature Publishing Group UK London}
}

@article{DQN,
  title={Human-level control through deep reinforcement learning},
  author={Mnih, Volodymyr and Kavukcuoglu, Koray and Silver, David and Rusu, Andrei A and Veness, Joel and Bellemare, Marc G and Graves, Alex and Riedmiller, Martin and Fidjeland, Andreas K and Ostrovski, Georg and others},
  journal={Nature},
  volume={518},
  number={7540},
  pages={529--533},
  year={2015},
  publisher={Nature Publishing Group}
}

@inproceedings{RL5,
  title={Scalable deep reinforcement learning for vision-based robotic manipulation},
  author={Kalashnikov, Dmitry and Irpan, Alex and Pastor, Peter and Ibarz, Julian and Herzog, Alexander and Jang, Eric and Quillen, Deirdre and Holly, Ethan and Kalakrishnan, Mrinal and Vanhoucke, Vincent and others},
  booktitle={Conference on Robot Learning},
  pages={651--673},
  year={2018},
  organization={PMLR}
}

@article{RRLforquad,
  title={Robust deep reinforcement learning for quadcopter control},
  author={Deshpande, Aditya M and Minai, Ali A and Kumar, Manish},
  journal={IFAC-PapersOnLine},
  volume={54},
  number={20},
  pages={90--95},
  year={2021},
  publisher={Elsevier}
}

@inproceedings{RARL,
  title={Robust adversarial reinforcement learning},
  author={Pinto, Lerrel and Davidson, James and Sukthankar, Rahul and Gupta, Abhinav},
  booktitle={International Conference on Machine Learning},
  pages={2817--2826},
  year={2017},
  organization={PMLR}
}

@inproceedings{ARDDPG,
  title={Action robust reinforcement learning and applications in continuous control},
  author={Tessler, Chen and Efroni, Yonathan and Mannor, Shie},
  booktitle={International Conference on Machine Learning},
  pages={6215--6224},
  year={2019},
  organization={PMLR}
}

@article{lee2025robust,
  title={Robust Deterministic Policy Gradient for Disturbance Attenuation and Its Application to Quadrotor Control},
  author={Lee, Taeho and Lee, Donghwan},
  journal={arXiv preprint arXiv:2502.21057},
  year={2025}
}

@article{RAP,
  title={Robust reinforcement learning using adversarial populations},
  author={Vinitsky, Eugene and Du, Yuqing and Parvate, Kanaad and Jang, Kathy and Abbeel, Pieter and Bayen, Alexandre},
  journal={arXiv preprint arXiv:2008.01825},
  year={2020}
}

@inproceedings{RARLineq,
  title={Robust adversarial reinforcement learning with dissipation inequation constraint},
  author={Zhai, Peng and Luo, Jie and Dong, Zhiyan and Zhang, Lihua and Wang, Shunli and Yang, Dingkang},
  booktitle={Proceedings of the AAAI Conference on Artificial Intelligence},
  pages={5431--5439},
  year={2022}
}

@book{Hbook,
  author		= "Ba{\c{s}}ar, Tamer and Bernhard, Pierre",
  title			= "H-infinity optimal control and related minimax design problems: a dynamic game approach",
  address		= "Boston",
  publisher		= "Springer Science \& Business Media",
  year			= "2008"
}

@article{HRL3,
  title={Learning {H}-infinity locomotion control},
  author={Long, Junfeng and Yu, Wenye and Li, Quanyi and Wang, Zirui and Lin, Dahua and Pang, Jiangmiao},
  journal={arXiv preprint arXiv:2404.14405},
  year={2024}
}

@article{OUnoise,
  title={On the theory of the Brownian motion},
  author={Uhlenbeck, George E and Ornstein, Leonard S},
  journal={Physical review},
  volume={36},
  number={5},
  pages={823},
  year={1930},
  publisher={APS}
}

@article{AdvRLSurvey,
  title={Robust deep reinforcement learning through adversarial attacks and training: A survey},
  author={Schott, Lucas and Delas, Josephine and Hajri, Hatem and Gherbi, Elies and Yaich, Reda and Boulahia-Cuppens, Nora and Cuppens, Frederic and Lamprier, Sylvain},
  journal={arXiv preprint arXiv:2403.00420},
  year={2024}
}

@article{SalvatoRealityGap,
  author={Salvato, Erica and Fenu, Gianfranco and Medvet, Eric and Pellegrino, Felice Andrea},
  title={Crossing the Reality Gap: A Survey on Sim-to-Real Transferability of Robot Controllers in Reinforcement Learning},
  journal={IEEE Access},
  year={2021}
}

@inproceedings{ZhaoSimToRealSurvey,
  author={Zhao, Wenshuai and Queralta, Jorge Peña and Westerlund, Tomi},
  title={Sim-to-Real Transfer in Deep Reinforcement Learning for Robotics: a Survey},
  booktitle={IEEE Symposium Series on Computational Intelligence (SSCI)},
  year={2020}
}

@article{GarciaFernandezSafeRL,
  author={Garc{\'i}a, Javier and Fern{\'a}ndez, Fernando},
  title={A Comprehensive Survey on Safe Reinforcement Learning},
  journal={Journal of Machine Learning Research},
  volume={16},
  number={1},
  pages={1437--1480},
  year={2015}
}

@inproceedings{Tobin2017DR,
  title={Domain randomization for transferring deep neural networks from simulation to the real world},
  author={Tobin, Josh and Fong, Rachel and Ray, Alex and Schneider, Jonas and Zaremba, Wojciech and Abbeel, Pieter},
  booktitle={2017 IEEE/RSJ International Conference on Intelligent Robots and Systems (IROS)},
  pages={23--30},
  year={2017},
  organization={IEEE}
}

\end{document}